# Automated dysgraphia detection by deep learning with SensoGrip

Mugdim Bublin[1*], Franz Werner[2], Andrea Kerschbaumer[2], Gernot Korak[3], Sebastian Geyer[3], Lena Rettinger[2], Erna Schönthaler[4], Matthias Schmid-Kietreiber[1]

[1]FH Campus Wien, University of Applied Sciences, Department Technics, Computer Science and Digital Communication, Vienna, Austria

[2]FH Campus Wien, University of Applied Sciences, Department Technics & Health Sciences, Health Assisting Engineering, Vienna, Austria

[3]FH Campus Wien, University of Applied Sciences, Department Technics, High Tech Manufacturing, Vienna, Austria

[4]FH Campus Wien, University of Applied Sciences, Department Health Sciences, Occupational Therapy, Vienna, Austria

**\*Corresponding author:**

Mugdim Bublin, PhD

FH Campus Wien, University of Applied Sciences,

Favoritenstrasse 226, 1100 Vienna, Austria

E-mail: mugdim.bublin@fh-campuswien.ac.at

# Abstract

Dysgraphia, a handwriting learning disability, has a serious negative impact on children's academic results, daily life and overall wellbeing. Early detection of dysgraphia allows for an early start of a targeted intervention. Several studies have investigated dysgraphia detection by machine learning algorithms using a digital tablet. However, these studies deployed classical machine learning algorithms with manual feature extraction and selection as well as binary classification: either dysgraphia or no dysgraphia. In this work, we investigated fine grading of handwriting capabilities by predicting SEMS score (between 0 and 12) with deep learning. Our approach provide accuracy more than 99% and root mean square error lower than one, with automatic instead of manual feature extraction and selection. Furthermore, we used smart pen called SensoGrip, a pen equipped with sensors to capture handwriting dynamics, instead of a tablet, enabling writing evaluation in more realistic scenarios.

# Introduction

Dysgraphia is a handwriting learning disability, which can have a serious negative impact on children's academic success and daily life [1], [2]. Cognitive, visual perceptual and fine motor skills are essential for learning to write [3]. Dysgraphia refers to the motor performance difficulties of handwriting, and therefore not the skills of spelling or text composition. Between 5 and 34% of children, never master legible, fluent and enduring handwriting, despite the appropriate learning over a period of 10 years [4], [5]. Early detection of dysgraphia is important to initiate a targeted intervention to improve the children's writing skills as a basis for a further academic career. Criterion-referenced assessment of handwriting skills in occupational therapy practice traditionally includes the quantitative and qualitative analysis of letter formation, spacing, alignment, legibility and handwriting speed [0].

In recent times several studies have investigated automatic dysgraphia detection by machine learning algorithms using a digital tablet [4], [7-12]. Tablet sensors enable collecting signals like the x and y coordinates, pressure, and tilt of the pen during writing. However, these studies deployed classical machine learning algorithms like Random Forest[4] and AdaBoost[7], which require manual feature extraction and selection.

Asselborn et al. [4] extracted 53 handwriting features. These 53 features are classified into four groups: static, kinematic, pressure and tilt features. Spectrum features seem to be the most important features: six out of eight features are related to frequency. The most discriminative four features were the Bandwidth of Tremor Frequencies and Bandwidth of Speed Frequencies (kinematic features), Mean Speed of Pressure Change (pressure feature) and space Between Worlds (static feature). Using these 53 features and the Random Forest classifier they achieved an F1-score of 97,98%. Accuracy is not reported due to imbalanced data set (56 children with dysgraphia and 242 typically developing children). In Asselborn et al. [5] PCA is used to reduce the number of relevant features and to enable a more data-driven classification.

Drotar & Dobes [8] extracted 133 features using mainly statistical measures leak min, mean, median and standard deviation of features like velocity, acceleration, jerk, pressure, attitude, azimuth, segment/vertical/horizontal length, pen lifts etc. These 133 features per task, merged together from all tasks produce 1176 features in total. From these 1176 features, 150 features are selected using weighted k-nearest neighbour feature selection. Finally, using these 150 features as input to the AdaBoost classifier, they obtained an accuracy of 79.5%. The data set was almost balanced: 57 children with dysgraphia and 63 normally developing children.

Dimauro et al. [9] used 13 pure text-based features like writing size, non-aligned left margin, skewed writing, insufficient space between words etc. and achieved an accuracy of 96% but with imbalanced data with 12 out of 104 children with dysgraphia.

On the other side, Devillaine et al. [10] used only graphical tablet sensor signals: x, y, and z positions and pressure data for feature extraction and no static features from text data. From the extracted features 10 features were selected by Linear SVM or Extra Trees. Finally, different classical machine learning algorithms were tested for classification of a balanced data set consisting of 43 children with dysgraphia and 43 typically developing children. The best performance: 73.4% accuracy was achieved with the Random Forest algorithm. Devillaine et al.[10] provided also a comparison of different classical machine learning algorithms for dysgraphia detection, with the highest F1-scored of 97,98% achieved by Asselborn et al. [4].

To summarize, all the cited papers used feature extraction and selection, which require high effort and expertise. Furthermore, the feature extraction is also subjective, based on expert´s experience and might miss some important features. In our project, we used an LSTM deep learning algorithm, which automatically extracts features from the raw sensor signals. To our knowledge, only a few research groups deployed deep learning for dysgraphia detection like Ghause et al. [11], achieving an F1-score of 98,16%, but this work used CNN with text images as input i.e. no dynamic features were used. Zolna et al. [12] achieved a significantly better performance in dysgraphia detection when using RNN (more than 90%) in comparison to CNN (25% - 39%), because RNN takes into account the dynamic aspects of the writing and CNN does not. However, they used only the trajectory of the consecutive points, as dynamic features and not pressures, velocity, accelerations and tilt as we did. They also used two layers LSTM but with 100 neurons in each layer and a drop-out layer with 50% drop probability.

In contrast to previous works, which used tablets, we used a pen equipped with sensors (SensoGrip) to capture handwriting dynamics: pressure, speed, acceleration and tilt, which looks and feels more like a real pen, and enables writing evaluation in more realistic scenarios. The SensoGrip System consists of the SensoGrip pen and an app designed for Android OS. The SensoGrip pen features an integrated microcontroller, which is able to communicate with the Android device via Bluetooth BLE. It also contains the necessary power supply, electronics and sensors for measuring tip and finger pressure as well as an IMU MEMS 3-axis accelerometer and 3-axis gyroscope. The microcontroller captures the pressure data as well as the data provided by the IMU and forwards it to an App on the Android device. The user is able to acquire feedback via built-in RGB LEDs or via the mobile app.

We achieved higher accuracy (more than 99%) than the cited works. Furthermore, our output is not binary: dysgraphia or no dysgraphia, but the scores of the German version of the Systematic Screening for Motor Handwriting difficulties (SEMS) [13]. The SEMS provides values between 0 and 12 for print and a maximum of 14 points for manuscript writing. More points indicate more difficulties in writing, and therefore these scores enable a finer granularity in handwriting evaluation.

In the sequel of this work, we present our results and discuss them, then we describe our methods and finally conclusions are provided.

## Methods

### The SensoGrip system

The core device for the acquirement of the data is the SensoGrip pen (see Figure 1), whose main components (1), (4) and (7) were manufactured using the additive manufacturing process of Hot Lithography. Apart from the outer shell (4) and the very tip of the pen (7), the light guide (6), which transports the signal light from the PCB (2) to the visible LED ring, was traditionally manufactured by milling and turning PMMA stock material. The light guide was then polished to make it fully transparent. The grip (3) was manufactured via indirect rapid prototyping using Hot Lithography to produce a split mold. Two-component silicone rubber was used to form the hollow cylindrical grip surfaces, which were then mounted on top of the corresponding finger sensors.

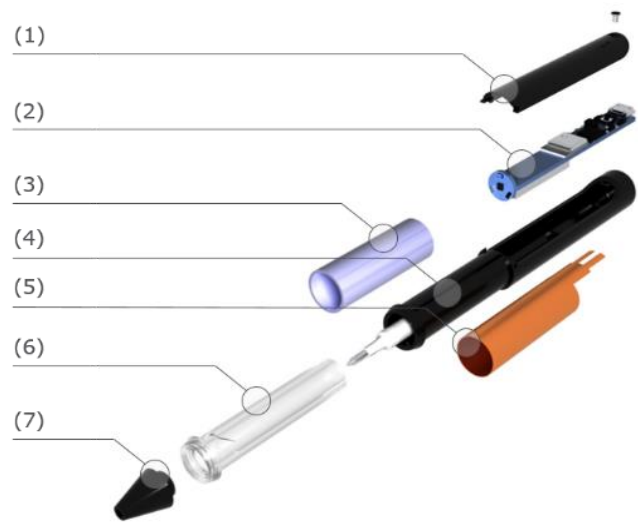

*Figure 1: SensoGrip structure*

The processing in the pen is done by a NINA B306 microcontroller unit. NINA-B3 series modules are small stand-alone Bluetooth 5 low energy microcontroller units (MCU) modules. For the measurement of the tip-pressure, a HSFPAR004A piezoresistive sensor is used. The finger pressure is captured with an FSR 406 force-sensing resistor. The MPU5060 IMU was chosen, due to low power consumption, low price, ease of use and a small form factor. It can deliver 3-axis accelerometer data as well as 3-axis gyroscope data. It also provides a built-in digital motion processor for sensor fusion. During the writing process, the different data streams provided by the sensors are captured by the SensoGrip pen and forwarded to the app on the Android device via BLE with corresponding time stamps.

**Data collection**

The data collection was conducted as part of a bachelor's thesis by two students of the Occupational Therapy Program at the University of Applied Sciences FH Campus Vienna Banhofer & Lehner, [14]. The aim was to collect writing data from children without impairments as part of a qualitative and quantitative study.

For this purpose 22 children aged 7-9 years were included in the study, of which 12 children were female and 10 male.

The children used the SensoGrip pen twice during the data collection and had the task of using the pen for at least five minutes to copy the sentences from the assessment "SEMS" (Vinçon, Blank & Jenetzky, [0]. Both tests were conducted on the same day, with a short break in between.

According to the regulations of the ethics commission of the FH Campus Wien, no ethics vote is required for such work. The development and evaluation of the test system itself was carried out prior to the aforementioned study in a comprehensive clinical trial with children in therapy in accordance with the Medical Devices Act under the number EK 21-042-0321. All participating children and their parents or legal guardians gave written informed consent before the study was conducted.

**Machine Learning Methods**

Our model consists of a deep learning Long Short Term Memory (LSTM) network for SEMS prediction using time series sensor values, and a Support Vector Machine (SVM) Classifier that combines SEMS prediction by LSTM with discreet values age and gender to predict the final SEMS score (see Figure 2).

We used LSTM networks for SEMS prediction since these networks can cope well with vanishing gradient problems typical for training recurrent networks [16]. For combining the output of the LSTM network with age and gender, we do not need deep learning, since we have only three discrete variables as input. Therefore, we used a well-known support vector machine (SVM) algorithm, which is among the best classical machine learning algorithms for such data [17].

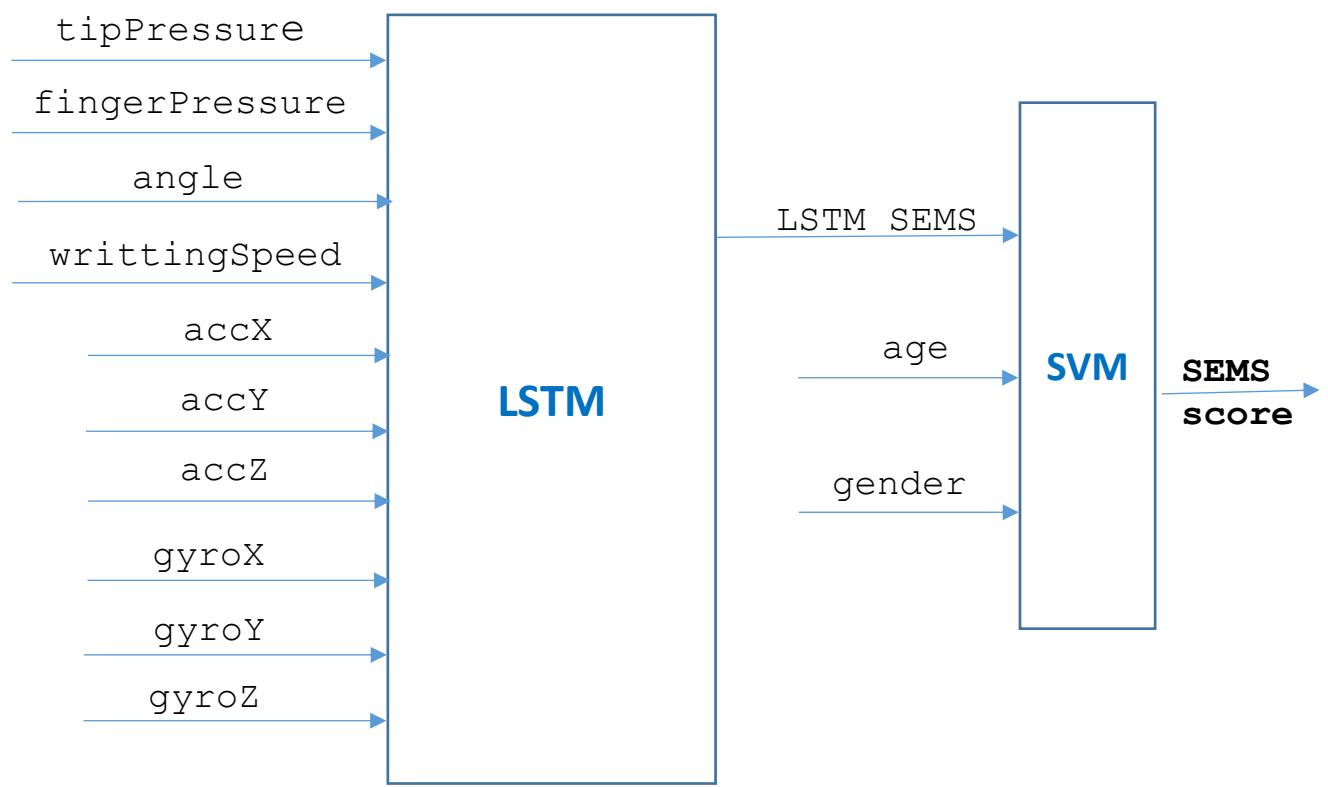

*Figure 2: Our model for SEMS prediction*

Mathematically speaking, we used a composition of two functions $f_{LSTM} : \mathbb{R}^{10 \times n} \rightarrow \mathbb{R}$ that predicts SEMS from the n values of 10 sensors using LSTM network (n is typically between 120 in our implementation, but can be further optimized) and $f_{SVM} : \mathbb{R}^3 \rightarrow \mathbb{R}$ that predicts the final SEMS score using SVM regression with the SEMS prediction from the LSTM network, age and gender as input features:

$$SEMS\_score = f_{SVM} \circ f_{LSTM} = f_{SVM}(f_{LSTM}(\boldsymbol{X}), age, gender)$$

Where $\boldsymbol{X}$ is a matrix containing the n time-series values of 10 sensors as depicted in Figure 2.

In order to increase the amount of training and test data, we divided the data file for each child into 20 separate time series and annotated them with the same SEMS score, which was estimated for the whole data file.

We used an LSTM network with two hidden layers, with 70 and 50 hidden units respectively. Between the hidden layers and after the second hidden layer, dropout layers were used to prevent overfitting (with a dropout rate of 20%). After the last hidden layer, a fully connected layer with one neuron was used producing the regression output. For optimization Adam optimizer algorithm was used with the initial learning rate of 0.005 and mean squared error as the performance measure. Optimization was done in 10 iterations per epoch and a total of 250 epochs.

For LSTM and SVM training, we divided the data in training, validation and test sets with ratios: 80%, 20% and 10% respectively. We used 10-fold cross-validations of the results i.e. the final results are obtained by averaging random trials, where in each trial we randomly selected 90% of data for training and 10% of data for the test.

In order to make our model interpretable, we applied the methods described in [18], based on viariable-wise hidden states with a mixture attention mechanism, to distinguish the contribution of each variable on SEMS score prediction (overall and for different time steps).

## Results

### Performance Measures

We used the Root Mean Squared Error (RMSE) between predicted and therapists' SEMS score as a basic measure for model evaluation:

$$RMSE = \sqrt{\frac{\sum (Predicted\_SEMS - Therapeut\_SEMS)^2}{Number\ of\ obeservations}})$$

In order to be able to compare our results with the results from the literature, we had to define a SEMS threshold above which the dysgraphia was detected since in the literature only binary classification: either dysgraphia or no dysgraphia is used. To that purpose, we defined a SEMS score at the level of 6 or above as an indication for handwriting difficulties, since with this threshold a good agreement is achieved with teachers' perceptions (98.21% specificity and 100% sensitivity)[0]. The threshold of 6 is estimated for grade 2 children with an average age of 7.6 years, which fits well with the average age of our children (7.2 years).

Consequently, accuracy and F1 score, are defined using following equations:

$$SEMS\ threshold\ for\ Dysgraphia\ detection: SEMS_{THR} = 7$$

$$True\ positives\ (TP) = \sum \left( (Predicted_{SEMS} \geq SEMS_{THR}\ AND\ Therapeut_{SEMS} \geq SEMS_{THR}) \right)$$

$$False\ positives\ (FP) = \sum \left( (Predicted_{SEMS} \geq SEMS_{THR}\ AND\ Therapeut_{SEMS} < SEMS_{THR}) \right)$$

$$True\ negatives\ (TN) = \sum \left( (Predicted_{SEMS} \geq SEMS_{THR}\ AND\ Therapeut_{SEMS} < SEMS_{THR}) \right)$$

$$False\ negatives\ FN = \sum \left( (Predicted_{SEMS} < SEMS_{THR}\ AND\ Therapeut_{SEMS} \geq SEMS_{THR}) \right)$$

$$Sensitivity = \frac{TP}{TP + FN}$$

$$Specificity = \frac{TN}{TN + FP}$$

$$Precision = \frac{TP}{TP + FP}$$

$$Recall = \frac{TP}{TP + FN}$$

$$Accuracy = \frac{TP + TN}{TP + TN + FP + FN}$$

$$F1_{score} = 2 * Precision * \frac{Recall}{Precision + Recall}$$

**Numerical Results and Discussion**

We optimized the LSTM network by using one, two or three hidden layers and different numbers of neurons in each layer respectively (see **Table 1**).

*Table 1: LSTM network optimization*

| Number of layers | Number of Hidden Units L1 | Number of Hidden Units L2 | Accuracy (%) | F1 Score (%) | RMSE |
|---|---|---|---|---|---|
| 1 | 80 | - | 98.29 | 71.87 | 0.97 |
| 1 | 100 | - | 98.00 | 63.65 | 1.07 |
| 1 | 120 | - | 98.22 | 67.82 | 1.04 |
| 2 | 70 | 40 | 99.18 | 86.97 | 0.99 |
| **2** | **70** | **50** | **99.44** | **92.31** | **0.89** |
| 2 | 80 | 50 | 99.35 | 94.01 | 0.87 |

As can be seen from Table 1, the best performance were achieved with the LSTM networks with two layers. Two layers seems to be a reasonably compromise between underfitting (one layer LSTM) and overfitting (three layers LSTM). According to our simulations, the optimal number of neurons in the first layer was 70 and in the second layer was 50 neurons, but further optimizations are possible, since we did not simulated many different combinations of neurons per layer.

In **Table 2** we provide performance comparisons of our model with other models. In most papers only accuracy is provided, whereas Asselborn et al. reports F1-Score since overall accuracy might be misleading if most of the children come from one group i.e. non-dysgraphia.

The accuracy is estimated under the assumption that SEMS score at the level of 6 or above is an indication for handwriting difficulties, according to Franken et al. [13].

*Table 2: Performance comparisons of our model with models from literature*

| Model | Accuracy mean +/- std (%) | F1-Score mean +/- std (%) | RMSE mean +/- std |
|---|---|---|---|
| Asselborn et al.[4] | - | 97.98 +/- 2.68 | - |
| Drotar & Dobes[6] | 79.50 +/- 3 | - | - |
| Dimauro et al[7] | 96 | - | - |
| Devillaine et al.[8] | 73.40 +/- 3.4 | - | - |
| Ghouse et al.[9] | 98.2 | 98.16 | - |
| Zolna et al.[0] | 90 | | |
| Our model | **99.54 +/- 0.57** | **95.98 +/- 4.57** | **0.85 +/- 0.11** |

As can be seen from **Table 2**, we achieved the highest accuracy in comparison to published results (over 99%). We obtained a lower F1-score than the F1-score published in 4, but we have also a high standard deviation of the F1-score (more than 4%), which is due to the relatively small sample size of the tested children.

We achieved an RMSE value below one, but RMSE was not reported in any of the cited papers. The other works made only binary classification decisions: dysgraphia or no dysgraphia, whereas we performed a regression of the SEMS score, which enables the evaluation of children's writing capabilities on the finer granularity scale between zero and twelve (the highest score obtained in our study was eight).

To distinguish the overall contribution of each variable on SEMS score prediction, we applied the variable-wise hidden states with a mixture attention mechanism method described in [18]. The impact of different variables on performance is presented in *Figure 3*.

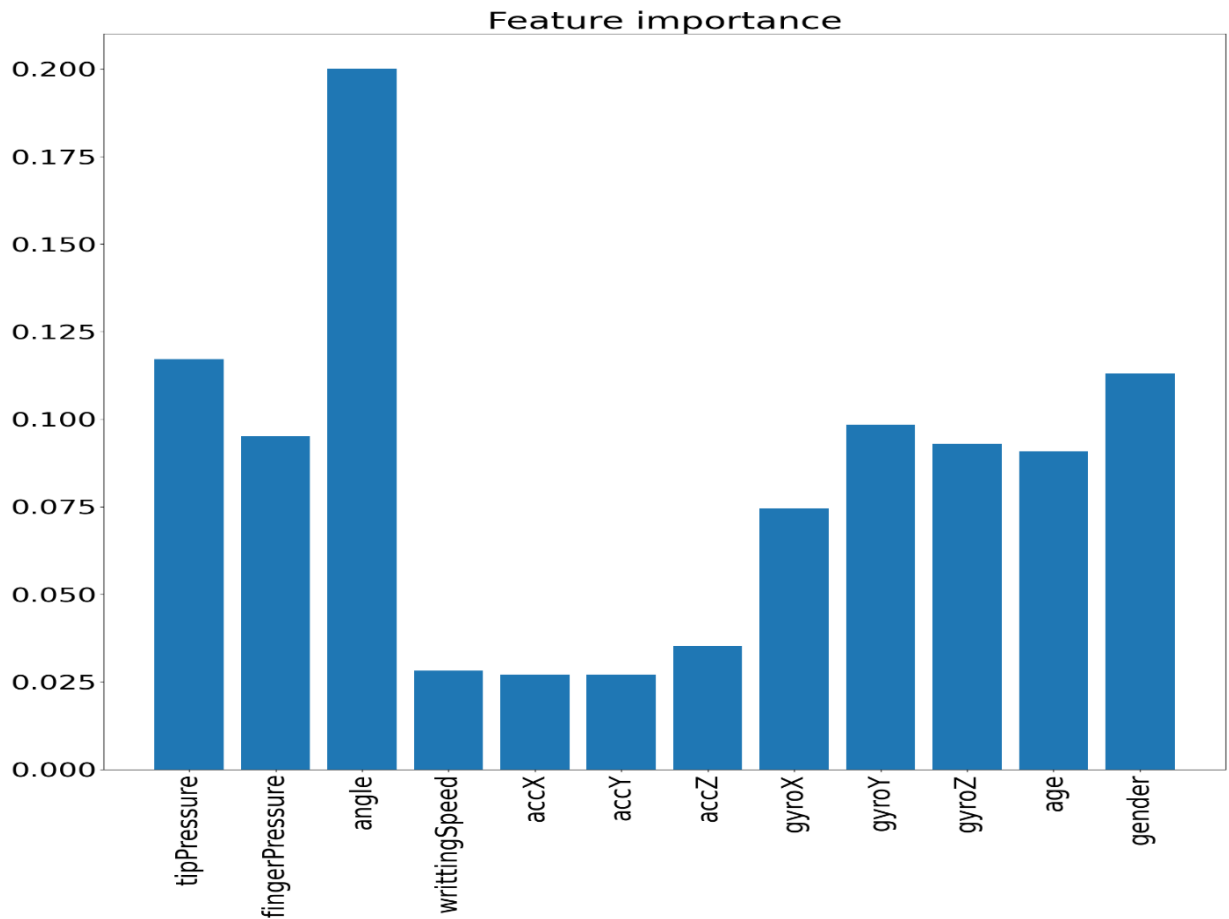

*Figure 3: Impact of different features (sensor values) on performance*

As can be seen from *Figure 3* the importance of different sensor values for the SEMS prediction performance is (in decreasing order): angle, age, finger pressure, gender, tip pressure, gyroZ, gyroY, gyroX, accZ, accY, accX und writingSpeed.

To identify more precisely what was wrong with writing in each time step, we also applied variable-wise hidden states with a mixture attention mechanism from [18]. On this way the children, therapist, teachers or parents can see what was wrong in each time step and possibly make appropriate correction (see

Figure 4).

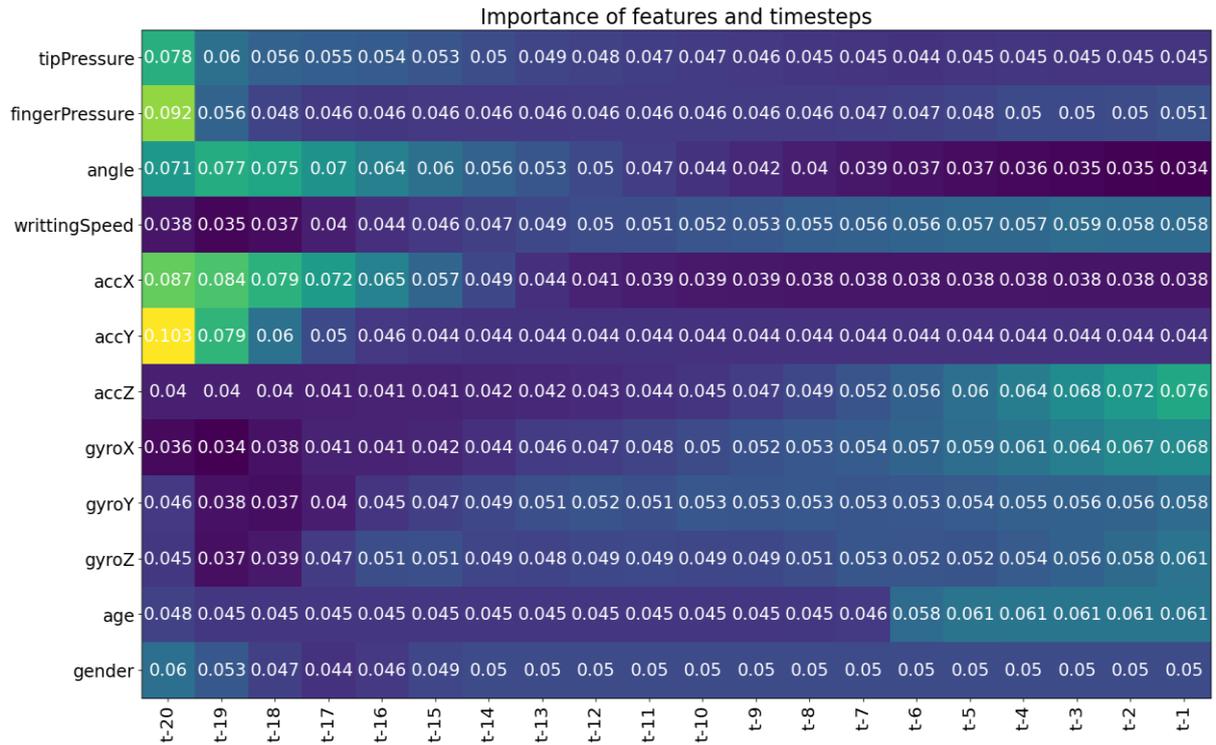

*Figure 4: Variable importance for different time steps*

In Figure 5 we made a comparison of the input signals and layer activations in the case of the best score with SEMS value of 0 (lefts) and the worst score with the SEMS value of 7 (rights).

As can be seen from Figure 5, the handwriting with the worse SEMS value had higher fluctuations in almost all input signals and caused higher activations of the LSTM hidden layers, especially the second and the last hidden layer.

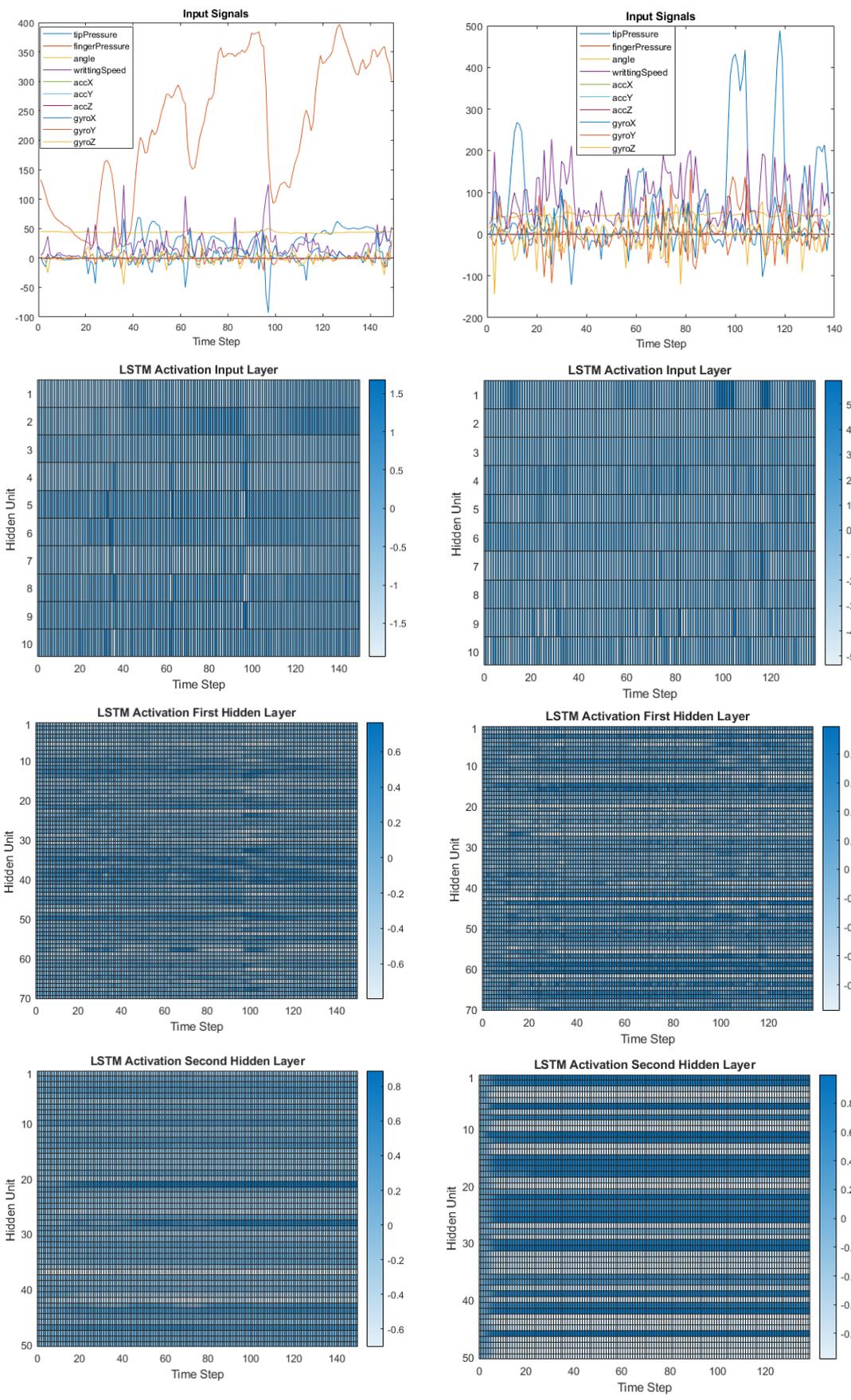

*Figure 5: Input signals (the first row) and layers' activations for the case of the best score: SEMS = 0 (lefts) and the worst score: SEMS = 7 (rights)*

## Conclusions

We developed a pen with sensor capabilities called SensoGrip that enables writing evaluation in realistic scenarios. Using a combination of deep learning (LSTM) and classical machine learning (SVM) approaches we were able to achieve low RMSE error in SEMS prediction (below 1) and high accuracy in detecting handwriting difficulties (above 99%). We did automatic instead of manual feature extraction and selection, which saves time and effort. In short, the advantages of our approach are high performance (accuracy over 99%, RMSE error lower than 1 on the SEMS scale between one and eight) and a more realistic scenario (using of a smart pen instead of a digital tablet).

In future works, collecting more data and further model optimization can bring even higher accuracy and lower RMSE. Furthermore, we are going to deploy our machine learning model on SensoGrip hardware to enable real time handwriting evaluation and provide a light feedback (red, yellow and green light) to children, teachers and parents. For future, we also plan to use SensoGrip for diagnosis and therapy of other disorders like Parkinson or Alzheimer deceases.

## Data Availability

Data and code are available as supplementary material on the publisher's website.

## Acknowledgements

This research was funded by the City of Vienna Magistrat 23, FH-Call 29, project 29-22: "Stiftungsprofessur – Artificial Intelligence" and FH – Call 30, project 30-25: "Artificial Intelligence and Virtual Reality Lab".

## Competing Interests Statement

No competing interests.